\pdfoutput=1

\documentclass[11pt]{article}

\usepackage{lipsum} 
\usepackage{stfloats} 
\usepackage[]{acl}
\usepackage{times}
\usepackage{latexsym}
\usepackage{hyperref}
\usepackage{amsmath}
\usepackage{amssymb}
\usepackage{array}
\usepackage{makecell}
\usepackage[T1]{fontenc}

\usepackage[utf8]{inputenc}

\usepackage{microtype}

\usepackage{inconsolata}

\usepackage{graphicx}
\usepackage{booktabs}
\usepackage{hyperref}
\usepackage{tabularx}

\usepackage{fontawesome5}

\usepackage{enumitem}
\setlist[itemize]{noitemsep, nolistsep}
\title{Optimizing Rare Word Accuracy in Direct Speech Translation with a Retrieval-and-Demonstration Approach}

\author{Siqi Li\textsuperscript{*1} \phantom{\and} Danni Liu\textsuperscript{*2} \phantom{\and} Jan Niehues\textsuperscript{2} \\
        \textsuperscript{1}University of California, Irvine, USA \\
        \textsuperscript{2}Karlsruhe Institute of Technology, Germany \\
        \texttt{siqil31@uci.edu}, \texttt{\{danni.liu, jan.niehues\}@kit.edu}
        }

\begin{document}
\maketitle

\begin{abstract}
Direct speech translation (ST) models often struggle with rare words. 
Incorrect translation of these words can have severe consequences,
impacting translation quality and user trust.
While rare word translation is inherently challenging for neural models due to sparse learning signals, 
real-world scenarios often allow access to translations of past recordings on similar topics.
To leverage these valuable resources, 
we propose a \textit{retrieval-and-demonstration} approach to enhance rare word translation accuracy in direct ST models.
First, we adapt existing ST models to incorporate retrieved examples for rare word translation,
which allows the model to benefit from prepended examples, similar to in-context learning.
We then develop a cross-modal (speech-to-speech, speech-to-text, text-to-text) retriever to locate suitable examples.
We demonstrate that standard ST models can be effectively adapted to leverage examples for rare word translation, 
improving rare word translation accuracy over the baseline by 17.6\% with gold examples and 8.5\% with retrieved examples.
Moreover, our speech-to-speech retrieval approach outperforms other modalities and exhibits higher robustness to unseen speakers.
Our code is publicly available\footnote{
\faGithubSquare:
\href{https://github.com/SiqiLii/Retrieve-and-Demonstration-ST}{\texttt{SiqiLii/Retrieve-and-Demonstration-ST}}
}.
\end{abstract}

\def\thefootnote{*}\footnotetext{Equal contribution; Siqi's work done while at KIT}\def\thefootnote{\arabic{footnote}}

\section{Introduction}
Speech translation (ST) traditionally involves cascading automatic speech recognition (ASR) and machine translation (MT) \cite{stentiford1988machine,waibel1991janus} to convert spoken language into text in a different language.
However, recent years have witnessed rapid progress in direct ST models \cite{anastasopoulos-etal-2021-findings,anastasopoulos-etal-2022-findings,agrawal-etal-2023-findings} that bypass intermediate text representations for lower inference latency and reduced error propagation \cite{sperber-paulik-2020-speech}.
Despite the advancements, 
accurately translating rare words like person names \cite{gaido-etal-2021-moby,10094689} remains a significant challenge for ST systems.
While infrequent, incorrect translations of rare words can severely degrade overall translation quality and even users' trust in the deployed models. 
Rare word translation is inherently difficult for ST models due to limited or absent learning signals.
Practically, however, valuable external resources hold the potential to address this issue. 
Real-world scenarios often allow access to translations from past recordings on similar topics, 
sometimes even from the same speaker. 
Similarly, human translators often leverage existing translations \cite{Bowker2005ProductivityVQ}, especially for special terminologies \cite{Brki2009UsingTM}.
Inspired by these observations,
we ask the question: 
How can we improve the rare word translation performance of direct ST models by leveraging an example pool that contains similar translations?

The envisioned approach faces challenges in both the \textit{retrieval} and \textit{translation} components.
First, the retrieval task is complicated by the variability of speech and the locality of rare words.
As the speaking condition for the same rare word differs in every utterance, 
source-side feature matching as often done in text translation \cite{zhang-etal-2018-guiding,bulte-tezcan-2019-neural,xu-etal-2020-boosting,cai-etal-2021-neural,hao-etal-2023-rethinking} 
is not sufficient to handle the pronunciation variations. 
Moreover, as rare words only constitute a small portion of the query and candidate utterances, 
the retriever must be able to locate the relevant information in long speech utterances.
For the translation model, 
integrating retrieved utterance-translation pairs is also non-trivial.
Standard models trained on sentence-level data require adaptation to ingest the examples.
Besides processing longer inputs, they also need to pinpoint both the acoustic features and corresponding textual translations of rare words.

\begin{figure*}[t]
  \includegraphics[width=16cm,trim={2.5cm 0 0 0},clip]{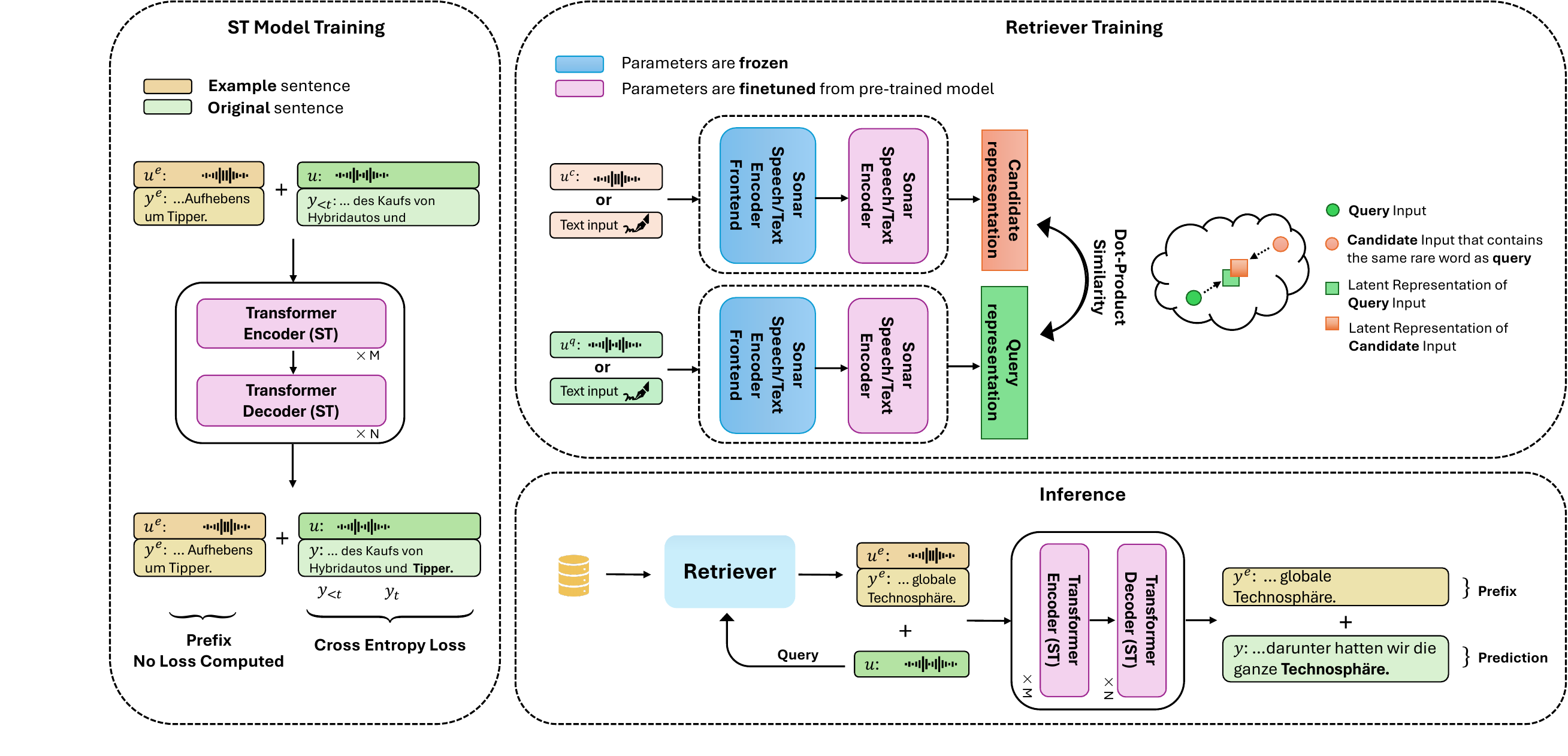}
  \caption{Proposed retrieval-and-demonstration framework: 
  At the ST model training stage (\S\ref{sec:ST Model}), example-prepended training data is used to instill in-context learning abilities in the S2T model. 
  At the retriever training stage (\S\ref{sec:example retrieval}), SONAR encoders are fine-tuned within the DPR architecture for our rare word task. 
  At the inference stage (\S\ref{sec:integration_st}), retrieved examples are used as demonstrations to facilitate the translation of rare words.}
  \label{fig:method}
\end{figure*}

Addressing the above challenges, 
we introduce a retrieval-and-demonstration framework (\autoref{fig:method}) effective for improving rare word translation accuracy of ST models. 
Specifically, we adapt standard ST models to benefit from prepended examples in a way similar to in-context learning \cite{brown2020language}, 
and then build a retriever to find suitable examples.
Building on recent multi-modal encoders \cite{duquenne2023sonar}, 
the retriever supports multiple modalities (speech$\rightarrow$speech, speech$\rightarrow$text, text$\rightarrow$text).
Second, we propose an evaluation methodology to adapt standard ST corpora, MuST-C \cite{di-gangi-etal-2019-must} in this case, for targeted assessment of rare words translation (\S\ref{sec:Data Construction}). Our main findings are:
\begin{itemize}[nolistsep,leftmargin=*]
    \item Standard direct ST models can be easily adapted to benefit from prepended examples for rare word translation, in a way similar to in-context learning (\S\ref{subsec:impact_of_demostration}). This improves rare word translation accuracy over the baseline by 17.6\% with gold examples and 8.5\% with retrieved examples.
    \item Text-to-text information retrieval architectures \cite{karpukhin-etal-2020-dense} can be effectively adapted for speech-based rare word retrieval, 
    yielding 33.3\% to 46.6\% top-1 retrieval accuracy under different modalities (\S\ref{subsec:retrieval_performance}).
    \item Compared to other modalities, speech-to-speech retrieval leads to higher overall translation quality and rare word translation accuracy (\S\ref{subsec:result_st_with_retrieval}), as well as more robustness to unseen speakers (\S\ref{subsec:unseen_speakers}).
\end{itemize}

\section{Proposed Framework}
\label{sec:proposed_framework}
Our retrieval-and-demonstration framework is illustrated in \autoref{fig:method}. 
First, a trained direct ST model is finetuned to ingest examples (\S\ref{sec:ST Model}), which serve as demonstrations of correctly translating the rare words in question.
During inference, 
given an utterance containing rare words, 
we retrieve (\S\ref{sec:example retrieval}) a relevant utterance and its translation as a demonstration to guide the inference (\S\ref{sec:integration_st}).



\subsection{Adapting ST Models to Ingest Examples}
\label{sec:ST Model}

\paragraph{Motivation}
Human translators often leverage example translations also known as \textit{translation memory} \cite{Bowker2005ProductivityVQ}, especially for domain-specific translation with terminologies \cite{Brki2009UsingTM}.
We aim to apply a similar approach to direct ST models.
The underlying idea mirrors that of in-context learning (ICL) \cite{brown2020language}, 
where providing models with task-specific examples during inference improves the quality of the generated output.
While ICL has been primarily observed on text-based LLMs \cite{brown2020language,min-etal-2022-rethinking,vilar-etal-2023-prompting},
we explore whether small- or medium-sized encoder-decoder-based speech translation models can also exhibit this capability.

\paragraph{Training}
To adapt standard ST models to ingest examples, 
the example utterance and translation must be included as context for training and inference.
An intuitive approach is to include the example as prefix in both input and output, 
as shown in the left side of \autoref{fig:method}.
This allows the output generation to be conditioned on the example utterance and translation as context.
Formally,
given an utterance $u$, 
let $\hat{y}$ be the target translation and $y$ the predicted translation. 
Let $(u^e,y^e)$ be an example utterance-translation pair.
We aim to adapt an ST model so that the model maximizes the probability of generating the correct translation $\hat{y}$, 
given the input utterance $u$ and example $(u^e,y^e): y =\operatorname*{arg\,max}_{\hat{y}} 
P(\hat{y}|u^e,y^e,u)$.
The difference to the standard training is that the example $(u^e,y^e)$ is included as context when generating the target translation.
For the training data,
for the $i$-th training utterance $u_i$,
an example utterance $u_i^e$ is prepended to it, forming a concatenated input
$u_i^e+u_i$.\footnote{Details on constructing the dataset is in \S\ref{sec:Data Construction}.}
The targets are also concatenated as $y_i^e +  \texttt{<SEP>} + y_i$, 
where $\texttt{<SEP>}$ is a special token indicating the separator between sentences.  
During training, the loss is only calculated on $y_i$ to prioritize the translation of the utterance after the example.\footnote{Including the loss on the prefix leads the finetuning step to end prematurely in preliminary experiments. The loss calculation is formally described in \autoref{appendix:masked_loss}.}
In doing so, we encourage the model to predict its outputs based on the context provided by the demonstration example.

\subsection{Example Retrieval}
\label{sec:example retrieval}
\paragraph{Formalization and Challenge} 
Given a query utterance $u$ containing a rare word $w$,
we aim to retrieve a relevant example $(u^e,y^e)$
from an example pool $\mathcal{D}=\{(u^1,y^1 ),…,(u^m,y^m )\}$
with a retrieval model $r$,
such that the rare word $w$ is spoken in utterance $u^e$.
Here $u^i$ indicates the $i$-th utterance and $y^i$ its translation.
As the query $u$ is only in speech, we face additional complexities compared to text-based retrieval. 
\textit{First}, speech is versatile, unlike text, which often has a standard writing system. 
The speaking condition for the same word varies in every recording, requiring a robust retriever that accounts for pronunciation variations.
\textit{Second}, speech sequences are magnitudes longer than text. The retriever must find fine-grained local features corresponding to the keywords in long sequences.
\textit{Third}, transcribing the query utterance first and then using text-based retrieval is suboptimal due to ASR errors, especially on rare words.

\paragraph{Architecture}
As the nature of our example retrieval task resembles information retrieval (IR) where relevant answers are retrieved given a question,
we take inspiration from IR approaches for our retriever.
In \textit{text-to-text} IR, 
a prominent architecture is the Dense Passage Retriever (DPR) \cite{karpukhin-etal-2020-dense}.
It has a \textit{dual-encoder} architecture, 
where one encoder encodes the questions, 
and the other encodes the passages potentially containing answers to the questions.
The retrieval model is trained with a contrastive objective, mapping question-passage (positive) pairs closer to each other in the latent space while pushing irrelevant (negative) pairs further apart.
During inference, passages closer to the encoded question by the dot-product similarity are returned as answers.
In our case, 
the utterances containing the same rare words are considered positive pairs, 
while those not sharing the same rare words are negative pairs.

\paragraph{Speech-to-Speech/Text Retrieval}
We propose to extend the DPR model to support querying from speech.
As the example utterances to be retrieved often also have text transcripts available, we consider the following retrieval modalities:
\begin{itemize}[nolistsep,leftmargin=*]
    \item Speech$\rightarrow$speech retrieval: we retrieve $u^e$ in speech using audio query $u$.
    \item Speech$\rightarrow$text retrieval: we retrieve $y^e$ directly using audio query $u$. This requires the retriever to support both modalities (text and speech). 
    \item Naïve text$\rightarrow$text retrieval: first transcribing the query utterance $u$ and then text-to-text retrieval for $y^e$. As discussed before, the risk of ASR errors especially on rare words renders this approach suboptimal. The additional inference time for running ASR makes it further unpractical.
\end{itemize}
Given these requirements, 
instead of initializing the dual encoders with pre-trained BERT \cite{devlin-etal-2019-bert} as in DPR \cite{karpukhin-etal-2020-dense}, 
we leverage recent speech-text joint representation models including SONAR \cite{duquenne2023sonar} and SpeechT5 \cite{ao-etal-2022-speecht5}. 


\subsection{Integrating Examples into ST Model}
\label{sec:integration_st}
\paragraph{Inference with Retrieved Examples}
During inference, the model is provided with a test input $u$ and a retrieved example $(u^e,y^e)$.
The example is prepended to test input in the same way as in training.
The example input-output pairs are integrated by forced decoding.
After the separator token (\texttt{<SEP>}), the model starts to autoregressively generate the output translation, conditioned additionally by the example utterance and translations.

\paragraph{Practical Considerations}
An advantage of our framework is its modularity.
The separation of the ST and retrieval modules 
enables straightforward upgrades to newer models in either component.
Moreover, the retrieval module can be implemented using highly optimized toolkits like FAISS \cite{johnson2019billion}, which ensures efficient retrieval without compromising inference speed.
Prepending examples however leads to increased inference latency as discussed in \S\ref{subsec:inference_latency}.

\begin{table}[ht!]
    \small
    \centering
    \setlength\tabcolsep{2pt}
    \begin{tabular}{m{2cm} c c c c}
    \toprule
    \textbf{Split}
    &\textbf{\# utt.} & \makecell{\textbf{Avg. utt.} \\ \textbf{duration (s)}} & \makecell{\textbf{Avg. \#} \\ \textbf{tokens}} & \makecell{\textbf{\# unique} \\ \textbf{rare words}}  \\
    \midrule
    train (original) &250942 &6.5 &27.1 &9512\\
    tst-COMMON
    &\phantom{0}\phantom{0}2580 &5.8 &25.3  &\phantom{0}157 \\
    \midrule
    rare-word pool 
    &\phantom{0}\phantom{0}9821 &9.7 &43.1    &8679 \\
    dev-rare-word
    &\phantom{0}\phantom{0}6932 &9.9 &42.8    &6244 \\
    tst-rare-word
    &\phantom{0}\phantom{0}2500 &9.9 &43.1    &2358 \\
    train-reduced
    &231689 &6.2 &25.8   &3164 \\
    \bottomrule
    \end{tabular}
    \caption{Dataset statistics.
    We split the original training set into the example pool with rare words (rare-word pool), dev/test sets for rare words (dev/tst-rare-word), and a reduced training set (train-reduced).
    The example pool simulates existing resources for querying.}
    \label{tab:data statistics}
\end{table}

\section{Experimental Setup}
\label{sec:Experimental Setup}

\subsection{Dataset Construction}
\label{sec:Data Construction}

For evaluation, we use the English-to-German subset of the \mbox{MuST-C} dataset \cite{di-gangi-etal-2019-must}, where the task is to translate from English public-speaking audio to German text.
To create a targeted test condition for rare words, 
we extract sentences containing rare words from the original training set to create dedicated sets.
The statistics of the original dataset and the newly created splits are in \autoref{tab:data statistics}.
The rare-word sets have higher average token counts due to: 1) longer utterance duration and 2) the rare words being segmented into finer-grained subwords.
Note that we only re-split the training set, 
leaving the official validation and test sets (tst-COMMON) unmodified.
Below we describe the dataset construction process in detail.

\paragraph{Rare Word Sets} 
Our data partition step is inspired by \citet{niehues-2021-continuous},
which re-splits parallel data based on word frequencies.
Specifically, 
from the English transcript, 
we find rare words by their \textit{corpus-level frequency}, 
choosing those appearing two or three times in the original training set. 
For rare words occurring twice, 
we move their corresponding utterances to the rare-word pool and the joint dev/tst set respectively, 
which creates a \textit{zero-shot} condition where the rare word is never seen in training.
For rare words occurring thrice, 
we follow the same strategy for two occurrences. 
The remaining third occurrence is retained in the reduced training set to create a \textit{one-shot} learning scenario,
where the rare word is seen once in the training set.
Finally, the aggregated dev/tst set is split into individual development and test sets for standard evaluation.
We analyze the rare word types in tst-rare-word by a named entity recognition (NER) model\footnote{\href{https://huggingface.co/urchade/gliner_large-v2.1}{Huggingface model} by \citet{DBLP:journals/corr/abs-2311-08526}} with results in \autoref{tab:NER analysis}.
A more detailed categorization of the words is in \autoref{appendix:rare_word_types}.

\begin{table}[ht!]
    \small
    \centering
    \setlength\tabcolsep{2pt}
    \begin{tabular}{c c c c c c} 
    \toprule
    \textbf{tst-rare-word}
    &\textbf{Person} &\textbf{Location} &\textbf{Tech} &\textbf{Food} &\textbf{Company} \\
    \midrule
    2358
    &130  &72   &29 &27 &25 \\
    \bottomrule
    \end{tabular}
    \caption{NER results on rare words in tst-rare-word with the number of unique words in each category.}
    \label{tab:NER analysis}
\end{table}    

\paragraph{Training Data with Prepended Examples} 
To adapt the ST model and to train the retriever, we need training data with prepended examples.
As most utterances lack rare words by the previously used corpus-level frequency (3164 rare words in 231k utterances in \autoref{tab:data statistics}), 
we train the retriever on \textit{simulated data} by treating words that have the lowest corpus-level frequency in each sentence as simulated rare words. 
Specifically,
we propose to use \textit{sentence-level} rare words to choose the prepended examples.
For each piece of the training data \((u^i,s^i,y^i)\), we identify the word \(w_s\) in \(s^i\) that has the least corpus-level frequency among all words in its transcript.
We then sample another training instance \((u^j,s^j,y^j)\) where \(s^j\) contains the same sentence-level rare word \(w_s\) as example.
In short, the retriever is trained without rare word retrieval data. 
In this \textit{zero-shot} training setup, 
the retrieval accuracy is limited by the strong mismatch between the train and test conditions.





\paragraph{Test Set with Gold Examples}
We also construct a variant of tst-rare-word set with gold examples, where the rare word in the test utterance is always present in the example.
This serves as an oracle condition for evaluating the ST model's ability to learn from perfect demonstrations.
As our data splitting procedure ensures that the rare words also occur in the example pool,
we select sentences from the rare-word pool containing the same rare words as those in the tst-rare-word set to serve as example sentences.
The example sentences are then prepended to test sentences in a way identical to that in the training set with prepended examples. 

\subsection{Model Configuration}
\paragraph{ST Model} 
We use the Transformer architecture \textsc{s2t\_transformer\_s} in \textsc{FairSeq S2T} \cite{wang-etal-2020-fairseq} for all our ST models.
To prevent the tokenizer from seeing the rare words during its  training,
which will cause an unfair test condition,
we train the SentencePiece \cite{kudo-richardson-2018-sentencepiece} tokenizer on the reduced train set after the utterances containing rare words are moved to dedicated splits (\autoref{tab:data statistics}).
Based on this vocabulary, 
we train the base model on the train-reduced set, 
closely following the hyperparameters from \citet{wang-etal-2020-fairseq}.
We then adapt the base model to ingest examples as described in \S\ref{sec:ST Model} using the reduced training set with prepended examples (\S\ref{sec:Data Construction}).
As the prefix tokens do not contribute to the overall loss  (\autoref{fig:method}), 
we double the effective batch size to keep the loss scale comparable to before.
Further details on training and inference are in \autoref{appendix:st_training_inference}.

\paragraph{Retriever} 
We use the DPR \cite{karpukhin-etal-2020-dense} architecture for the retriever.
The encoders are initialized with either SONAR \cite{duquenne2023sonar} or SpeechT5 \cite{ao-etal-2022-speecht5}.
For both models, we use the encoder only and discard the decoder.
DPR requires fixed-size embeddings from its encoders.
For SpeechT5, we mean-pool over the sequence length.
For SONAR, we use the built-in attention-pooling for the speech encoder and mean-pooling for the text encoder.
The dual encoders in DPR are trained on the reduced training set with prepended examples.
Each sentence's example serves as a positive example, 
while examples from other sentences in the batch are in-batch negatives.
Only the top layer of the encoders is trained, 
as the lower layers of the encoders are likely responsible for extracting low-level acoustic features. 
These features are considered less relevant for our retrieval task, which focuses on word-level information. 
Another reason is memory efficiency in training.
Further details on training and inference are in \autoref{appendix:retrieval_training_inference}.

\subsection{Evaluation}
\paragraph{Metrics} 
We evaluate speech translation quality with BLEU \cite{papineni-etal-2002-bleu}\footnote{sacreBLEU \cite{post-2018-call} signature:\newline nrefs:1|case:mixed|eff:no|tok:13a|smooth:exp|version:2.4.2}
and COMET \cite{rei-etal-2020-comet}\footnote{with \texttt{Unbabel/wmt22-comet-da}; $\times$100 for readability. The COMET models take text transcripts as source.}.
For the accuracy of rare word translation, 
we evaluate how many unique lemmatized rare words in the test set are translated. We use the spaCy toolkit \cite{Honnibal_spaCy_Industrial-strength_Natural_2020} for word lemmatization and used AWESoME Aligner \cite{dou-neubig-2021-word} for en-de word-level alignment.
For rare word accuracy, 
we further distinguish between rare words appearing once or never appear in the training set (\S\ref{sec:Data Construction}), 
which corresponds to the \textit{one-shot} and \textit{zero-shot} accuracy. 
For the retriever, we use top-1 retrieval accuracy to evaluate the retriever's performance.
Only the first retrieved examples are used as demonstrations in the ST model. 

\paragraph{Evaluation Difficulty}
As described in \S\ref{sec:Data Construction}, our rare word sets are based on rare words from the \textit{source-side} English transcripts.\footnote{ 
Constructing these sets based on target-side rare words would be unrealistic since the target is unavailable in practice.}
Due to the flexibility of translation, even with gold examples, some rare words are translated differently in the example translation versus the reference translation of the actual test sentence.
Only 845 of the 2500 unique words are translated to identical target words when using gold examples. 
Therefore, the highest possible accuracy is 33.8\% given this strict evaluation.\footnote{Ideally, beyond lexical matches, synonyms and other alternative translations should also be considered. As the evaluation of these cases is non-straightforward, we choose the strict lexical evaluation.}


\section{Main Results}
Before presenting the results of our proposed framework, 
we confirm that our baseline model performs on par with those reported in the literature.
The details are in \autoref{appendix:comparison}.

\subsection{Impact of Demonstration}
\label{subsec:impact_of_demostration}

    
    
\paragraph{Direct ST models can effectively learn from demonstration at inference time.}
To independently analyze the ST model's ability to learn from the prepended examples, 
we first assume an oracle retrieval model by using gold examples which always contain the rare words in question.
The results are in row (2) of \autoref{tab:rareword_translation_accuracy}.
Compared to the baseline in row (1),
this model
achieves substantially higher overall rare word translation accuracy ($+$17.6\% abs.), 
with a larger gain in zero-shot ($+$18.8\%) than one-shot accuracy ($+$15.3\%). 
Nonetheless, this gain comes at the cost of overall translation quality ($-$0.2 BLEU, $-$2.3 COMET).
A potential reason is that the prepended example sentences make the input sequences much longer and therefore create more difficulty for learning.
Nonetheless, 
since rare words are often important named entities, capturing them correctly is as crucial if not more than the overall translation quality scores.
Overall, the results suggest that task-specific demonstrations provided at inference time can effectively enhance rare word translation accuracy of direct ST models.


\paragraph{Quality of the given demonstration matters.} 
Next, we study the impact of the demonstration quality.
In contrast to the gold examples before, we now use random examples that do not contain rare words relevant to the sentence to be translated. 
The results are in row (3) of \autoref{tab:rareword_translation_accuracy}. 
This led to a decline in translation quality ($-$1.3 BLEU, $-$2.4 COMET) and rare word accuracy.
These results indicate that irrelevant demonstrations are harmful.

\paragraph{Seeing rare words only in training does not sufficiently improve their translation accuracy.}
Instead of retrieving data from the rare-word pool as demonstration, 
a simple alternative is to add these data in training.
Here, we add the rare-word pool into the training set and train an identical model to the baseline.
The results are in row (4) of \autoref{tab:rareword_translation_accuracy}.
Overall, the rare word accuracy only sees a slight increase compared to row (1), 
with an absolute accuracy improvement of 3.7\%, which is far less than using gold example sentences (+17.6\% overall).
This indicates that training with rare words alone is insufficient for improving their translation accuracy.
This is likely because of the limited training signal for rare words,
as each appears only once or twice.
Note that the translation quality scores under this data condition also improved, which is likely a result of the additional training data.
\begin{table*}[t]
    \small
    \centering
    \begin{tabular}{l c c c c c}  
    \toprule
    \textbf{ST Model} 
    & \textbf{BLEU} & \textbf{COMET} & \makecell{\textbf{Overall} \\ \textbf{acc (\%)}} & \makecell{\textbf{0-shot} \\ \textbf{acc (\%)}} & \makecell{\textbf{1-shot} \\ \textbf{acc (\%)}} \\
    
    \midrule
    (1) baseline model (on train-reduced)
    &17.2 &57.9  &11.8 &11.0  &13.3  \\
    (2) adapted + gold example
    &17.0  & 55.6  &\textbf{29.4} &\textbf{29.8}  &\textbf{28.6}  \\
    (3) adapted + random example
    &15.7 & 53.2 & \phantom{0}8.8 & \phantom{0}8.4  & \phantom{0}9.7  \\
    (4) train on \{train-reduced + rare-word pool\} (more data)
    &\textbf{17.9}  &\textbf{59.0} &15.5 &14.7  &17.2  \\
    \midrule
    \textbf{Using retrieved examples} \\
    (5) adapted + text (gold transcript)$\rightarrow$text 
    &15.2 & 54.4 &20.1 &19.6  &\textbf{21.2} \\
    (6) adapted + speech$\rightarrow$text
    &15.3   &  54.0 &18.8 &18.2  &20.2 \\
    (7) adapted + speech$\rightarrow$speech
    &\textbf{16.2} &\textbf{55.3}  &\textbf{20.3} &\textbf{20.3}  &20.2 \\
    
    \bottomrule
    \end{tabular}
    
    \caption{
    Translation quality (BLEU$\uparrow$, COMET$\uparrow$) and rare word accuracy$\uparrow$ (overall, 0- and 1-shot) of different models on the tst-rare-word split. 
    The lower section uses retrieved examples from the retriever (\S\ref{subsec:result_st_with_retrieval}).
}
\label{tab:rareword_translation_accuracy}
\end{table*}

\begin{table}[t]
    \small
    \centering
    \setlength\tabcolsep{3.5pt}
    \begin{tabular}{l c c c}
    \toprule
    \textbf{Retrieval Model}
    & \textbf{T\textrightarrow{}T} 
    & \textbf{S\textrightarrow{}T} 
    & \textbf{S\textrightarrow{}S} 
     \\
    \midrule
    (1) Orig. DPR w/ BERT (pretrained)
    & \phantom{0}2.0  &$-$  &$-$    \\
    (2) Orig. DPR w/ BERT (finetuned)
    &\textbf{55.8}  &$-$  &$-$    \\
    (3) DPR w/ SpeechT5 (finetuned)
    & \phantom{0}0.1  &0.0  & \phantom{0}0.0    \\
    (4) DPR w/ SONAR (pretrained)
    &28.7  &22.3  &20.6   \\
    (5) DPR w/ SONAR (finetuned) 
    &46.6  &\textbf{33.3}  &\textbf{41.3}   \\
    
    \bottomrule
    \end{tabular}
    \caption{Top-1 retrieval accuracy (\%) of different retrievers on 3 modalities of text-to-text (T$\rightarrow$T), speech-to-text (S$\rightarrow$T), and speech-to-speech (S$\rightarrow$S) on the tst-rare-word split. 
    T$\rightarrow$T retrieval uses gold transcripts as query.}
    \label{tab:retrieval_performance}
\end{table}
\subsection{Retrieval Performance}
\label{subsec:retrieval_performance}
Before integrating retrieved examples into the ST model, 
we analyze the retrieval performance alone with results in \autoref{tab:retrieval_performance}.
To establish the upper bounds of retrieval performance, we first use the original DPR model for text-to-text retrieval with gold transcripts of the query utterances and examples. 
As shown in row (1) of \autoref{tab:retrieval_performance}, directly using the pretrained DPR for QA is not sufficient for our task of rare word retrieval. 
Fine-tuning DPR’s encoders (row (2)) on our task enables effective rare word retrieval in a text-to-text setting (55.8\%). 

\paragraph{Encoder choice is crucial for successful retrieval.}
We proceed by adapting the original DPR to retrieval from speech. 
Overall, we notice that the choice of the encoder heavily impacts the retrieval performance. 
With SONAR, using the pretrained encoders already achieves partial success in fulfilling the task (row (4) in \autoref{tab:retrieval_performance}), with finetuning further improving the results (row (5)).
However, finetuning SpeechT5 proves insufficient for learning the task (row (3)).
We believe that the discrepancy primarily arises from the models' ability to aggregate information over the sentence length: 
SONAR is explicitly trained to aggregate it into fixed-size embeddings while SpeechT5 lacks such a mechanism.
Naïve mean-pooling over sequence length fails to create meaningful embeddings over long sequences like speech, as well as character-level text representations used in SpeechT5.



\paragraph{Speech$\rightarrow$speech outperforms speech$\rightarrow$text retrieval.}
While we initially expected speech-to-speech retrieval to be more challenging than speech-to-text retrieval due to the high variability of speech, the finetuned retriever in (5) of \autoref{tab:retrieval_performance} shows stronger performance on speech$\rightarrow$speech retrieval than speech$\rightarrow$text (41.3\% vs. 33.3\%). 
We suppose that the reason is the modality gap between text and speech, which makes it more challenging to bridge the two different types of data.

\subsection{ST Performance with Retrieved Examples}
\label{subsec:result_st_with_retrieval}
\paragraph{Correlation between retrieval accuracy and translation quality:}
As the retriever based on finetuned SONAR showed the most promising retrieval results (\autoref{tab:retrieval_performance}), 
we use the retrieved examples from this model to guide the ST. 
The results are in rows (5), (6), and (7) of \autoref{tab:rareword_translation_accuracy}. 
When comparing the performance of the three retrieval modalities, 
retrieval accuracy does not always translate to improved overall translation quality or rare word accuracy.
Although text-to-text retrieval using gold transcripts had the highest retrieval accuracy (\autoref{tab:retrieval_performance}), 
its integration into the ST model resulted in lower translation quality compared to speech-to-speech retrieval. 
Moreover, in practice, we still need an ASR model to derive the transcripts that likely contain errors, especially on rare words. 
This introduces additional limitations to the text-to-text retrieval approach.
Overall, these results show that speech-speech retrieval is more effective than the other modalities in improving rare word translation accuracy.
Despite the improvement in rare word translation accuracy, 
we also note the drop in translation quality compared to the baseline (row (7) vs. (1); $-$1.0 BLEU and $-$2.6 COMET).
We expect that increasing the robustness of the ST model to examples containing incorrect rare words, for instance by including such examples in training, 
could mitigate this negative impact.




\paragraph{Does speech$\rightarrow$speech retrieval help by implicit speaker adaptation?} 
Speech-to-speech retrieval could be particularly effective in finding same-speaker utterances due to the access to acoustic information. 
This raises the hypothesis that if the prepended example originates from the same speaker as the utterance to be translated, 
translation quality could be improved by implicit speaker adaptation \cite{DBLP:conf/asru/SaonSNP13},
where the model benefits from adapting to the specific speaker's voice characteristics.
To test this, we analyze the proportion of retrieved sentences from the same speaker across different retrieval modalities. 
The results in \autoref{tab:retrieval_speaker_analyze} show similar percentages for all three scenarios, 
indicating that the gains by speech-to-speech retrieval do not stem from speaker adaptation.
\begin{table}[ht!]
    \small
    \centering
    \setlength\tabcolsep{4pt}
    \begin{tabular}{l c c c}
    \toprule
    \textbf{DRP + SONAR finetuned}
    & \makecell{\textbf{T\textrightarrow{}T}} 
    & \makecell{\textbf{S\textrightarrow{}T}}
    & \makecell{\textbf{S\textrightarrow{}S}}  \\
    \midrule
    Examples from same speaker (\%)
    &50.3  &53.4    &50.2 \\
    \bottomrule
    \end{tabular}
    \caption{
    Proportion of retrieved examples from the same speaker as the utterance to be translated for the three retrieval modalities on tst-rare-word. 
    }
    \label{tab:retrieval_speaker_analyze}
\end{table}    

\section{Further Analyses and Discussions}
\subsection{Effects on Unseen Speakers}
\label{subsec:unseen_speakers}
  
Now we push the approach further under the challenging scenario of unseen speakers, i.e., the example pool does not contain any utterance from the speaker of the test utterance.
Specifically, during retrieval, we ignore utterances from the same speaker as the query utterance.
As shown in \autoref{tab:retrieval_speaker_trans_acc}, 
this harms retrieval accuracy substantially, losing 14.9\% to 23.4\% compared to \autoref{tab:retrieval_performance} for the three modalities.
This is mainly due to the limited coverage of the rare-word pool, 
which contains only one sentence for most rare words.
Excluding the speaker also excludes the rare word.
However, 
the BLEU scores and overall rare word translation accuracy change only slightly compared to \autoref{tab:rareword_translation_accuracy}:
T$\rightarrow$T ($-$0.6 BLEU, $-$1.5\%), 
S$\rightarrow$T ($-$0.3 BLEU, $-$3.2\%), 
S$\rightarrow$S ($+$0.2 BLEU, $-$1.0\%).  
This demonstrates that our approach, especially when using speech$\rightarrow$speech retrieval, 
is relatively robust to unseen speakers.

\begin{table}[ht!]
    \small
    \centering
    \setlength\tabcolsep{2pt}
    \begin{tabular}{l c c c c c}
    \toprule
    \makecell[l]{\textbf{Retrieval} \\ \textbf{modality}}
    &\makecell{\textbf{Retrieval} \\ \textbf{acc (\%)}}
    &\textbf{BLEU} 
    &\makecell{\textbf{Overall} \\ \textbf{acc (\%)}} & \makecell{\textbf{0-shot} \\ \textbf{acc (\%)}} & \makecell{\textbf{1-shot} \\ \textbf{acc (\%)}} \\
    
    \midrule
    (5) T$\rightarrow$T
    &23.2  &14.6 &18.6    &18.5 &18.7\\
    (6) S$\rightarrow$T
    &18.4 &15.0 &15.6 &15.6    &15.7\\
    (7) S$\rightarrow$S
     &\textbf{23.5} &\textbf{16.4} &\textbf{19.3}    &\textbf{18.8} &\textbf{20.2}\\
    \bottomrule
    \end{tabular}
    \caption{
    Retrieval and ST performance on unseen speakers. 
    Compared to \autoref{tab:rareword_translation_accuracy}, 
    S$\rightarrow$S retrieval has the least decrease in translation quality and rare word accuracy.
    }
    \label{tab:retrieval_speaker_trans_acc}
\end{table}  

\subsection{Qualitative Example}
\autoref{tab:translation example} shows an example where our approach creates partially correct translation for the named entities ``Patrice and Patee''.
To avoid cherry-picked results, we include more examples where our approach succeeds and fails in \autoref{appendix:sentence_examples}.

\begin{table}[h]
	\centering
	\small 
	\begin{tabularx}{\columnwidth}{X}
	\toprule
		\textbf{Source} (transcript): {\color{violet}Patrice and Patee} set out most days to go out hunting in the forest around their homes.
            \\
		\textbf{Baseline} (\autoref{tab:rareword_translation_accuracy} row $(1)$): Die {\color{purple}Bäume und Petes} {\color{gray}(Trees and Petes)} setzten die meisten Tage hinaus, um in den Wäldern um ihre Häuser zu pumpen.
		\\
        \textbf{Adding rare-word pool to training} (\autoref{tab:rareword_translation_accuracy} row $(4)$): {\color{purple}Patrizinpathie} {\color{gray}(Patrizinpathie)} setzte sich in den meisten Tagen um die Jagd in den Wäldern um ihre Häuser.
        \\
        \textbf{Speech$\rightarrow$speech example} (\autoref{tab:retrieval_performance} row $(5)$): Sie heißen {\color{blue}Patrice und Patee} {\color{gray} (Their names are Patrice and Patee.)}.
        \\
        \textbf{Adapted ST + speech$\rightarrow$speech} (\autoref{tab:rareword_translation_accuracy} row $(7)$): {\color{blue}Patrice und Patee}{\color{purple}tee} setzten die meisten Tage, um in den Wäldern um ihre Häuser herum jagen zu können.
        \\
        \textbf{Target}: {\color{blue}Patrice und Patee} {\color{gray} (Patrice and Patee)} gehen fast jeden Tag jagen in dem Wald rundum ihr Heim.\\
	\bottomrule
	\end{tabularx}
	\caption{An example of our retrieval-and-demonstration approach improving the translation of rare words.}\label{tab:example_knn}
    \label{tab:translation example}
\end{table}

\subsection{Analyses of Retrieval Performance}
\begin{figure}[ht!]
  \includegraphics[width=\columnwidth]{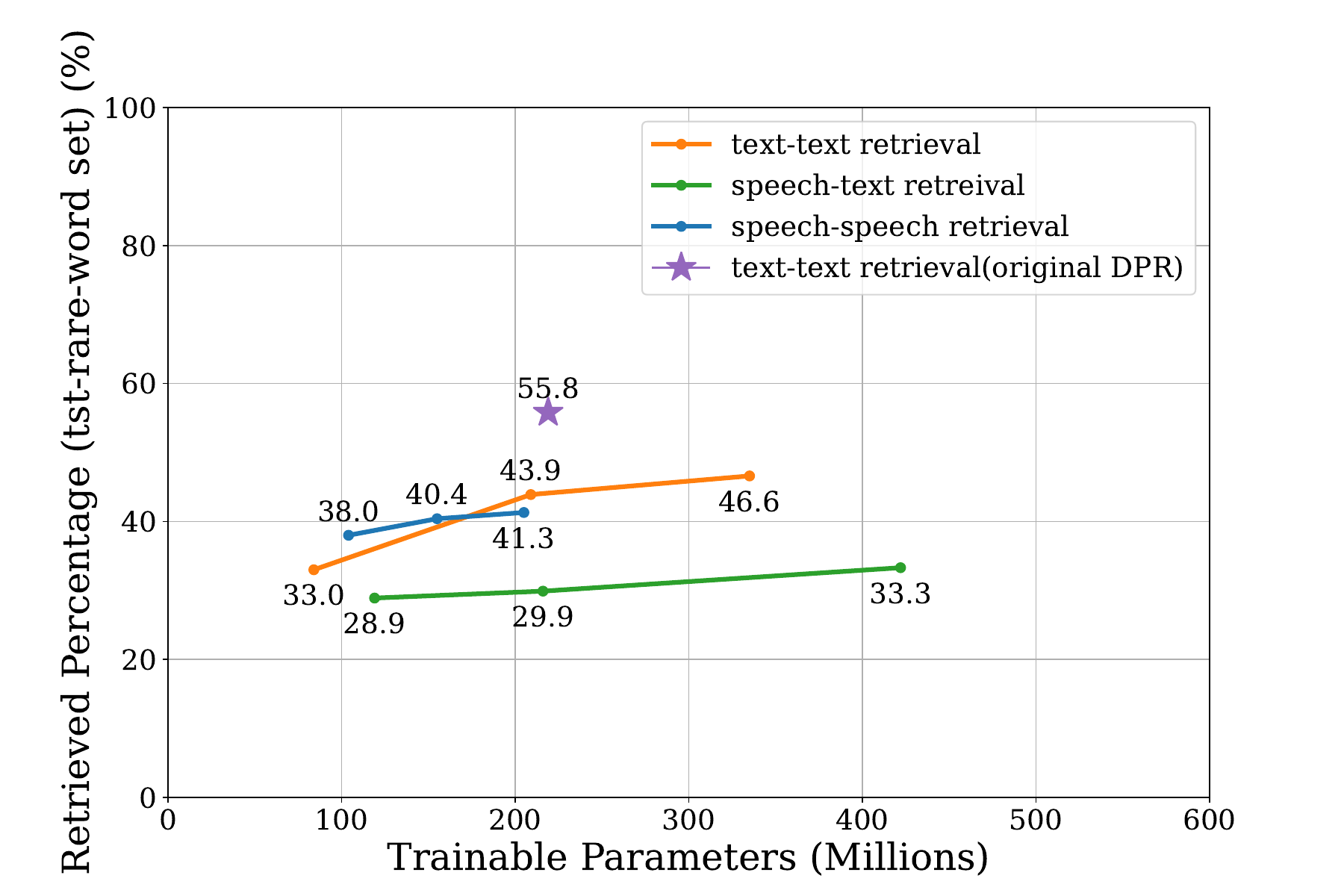}
  \caption{Retrieval performance of the SONAR-based retriever for different numbers of trainable parameters.}
  \label{fig:retrieve_experiments}
\end{figure}

In our main experiments, we partially finetuned the DPR encoders.
We now investigate the impact of different numbers of trainable parameters in the retriever.
As shown in \autoref{fig:retrieve_experiments}, 
the retrieval performance of the SONAR-based retriever is stable across 100 to 500M trainable parameters out of a total of over 1.3B parameters. 
This indicates that the retriever can maintain nearly consistent performance despite changes in model capacity.

\subsection{Potential of Using More Examples}
Few-shot learning is more often performant than one-shot learning because it provides the model with a broader context and more varied examples. 
However, as shown in \autoref{tab:retrieve_top10}, the increase in retrieval accuracy with additional top-10 examples is still not substantial compared to the top-1 result.  
Including multiple examples also makes input sequences significantly longer, 
especially as audio inputs are factors longer than text.
This not only poses a challenge for the model
but would also significantly slow down the inference speed, which we aim to avoid.
For these reasons, we do not further explore the potential of using more examples.

\begin{table}[ht!]
    \small
    \centering
    \setlength\tabcolsep{4pt}
    \begin{tabular}{l c c c}
    \toprule
    \textbf{DPR + SONAR ft.}
    & \makecell{\textbf{T\textrightarrow{}T}}  
    & \makecell{\textbf{S\textrightarrow{}T}}  
    & \makecell{\textbf{S\textrightarrow{}S}} \\ 
    \midrule
    Top 1
    &46.6  &33.3    &41.3 \\
    Top 5
    &60.4  &48.0    &56.2 \\
    Top 10
    &64.6  &53.1    &61.1 \\
    \bottomrule
    \end{tabular}
    \caption{
    Top-10 retrieval performance (\%) of the SONAR-based retriever on the tst-rare-word set.
    }
    \label{tab:retrieve_top10}
\end{table}    

\subsection{Inference Latency of Including Examples}
\label{subsec:inference_latency}
A downside of our approach is the additional inference latency due to longer prefixes, as inherent in other vanilla in-context learning approaches.
On the same GPU (NVIDIA Titan RTX) with batch size 1, the average inference time is 0.35s per sentence (system in Row 1, \autoref{tab:rareword_translation_accuracy}) and 0.82s after adding examples (average of the systems in Row 2-7, \autoref{tab:rareword_translation_accuracy}). 
The main contributor of the additional latency is the roughly doubled input sequence length. 
The text prefixes from the prepended examples are incorporated by forced decoding and do not incur much latency.

\subsection{Potential of Using Chunk-Based Examples}
Our in-context examples are in the form of parallel data.
An alternative is to use chunks instead of unprocessed parallel data.
In this case,
as the source and target of the in-context examples have to be aligned, creating the chunk-based example pool requires two additional alignment steps: 
audio-transcript alignment and transcript-translation alignment. 
While both steps have established off-the-shelf tools, 
this significantly complicates the workflow. 
Increasing the number of retrieval candidates may also increase the difficulty of the retrieval task. 
A main advantage of using chunks is the reduced inference latency as the prefixes are shorter.
Moreover, shorter context may be easier for the model to locate and utilize.
We leave the exploration of this alternative for future work.

\subsection{Reusing ST Encoder for Retrieval}
In our main experiments, we use SONAR for retrieval.
An attractive alternative is to use the encoder of pretrained ST models for retrieval, which would dramatically reduce the total model size at inference.
However, based on our comparison to using the SpeechT5 encoder for retrieval (Row 3, \autoref{tab:retrieval_performance}) and interpretations, 
models that do not explicitly shrink the sequence length dimension into more compact representations are likely unable to perform the retrieval task. 
Therefore, we believe the encoders of existing ST models would need to learn an aggregation mechanism like in SONAR to be ready for the retrieval task.

\section{Related Work}

\paragraph{Retrieval-Augmented Translation}
Our work falls within the paradigm of retrieval-augmented translation (RAT) \cite{simard-langlais-2001-sub,koehn-senellart-2010-convergence,tu-etal-2018-learning,DBLP:conf/iclr/KhandelwalFJZL21},
which augments a translation model with results retrieved from a translation memory.
Prior works on RAT primarily focus on text-to-text translation \cite{zhang-etal-2018-guiding,DBLP:conf/aaai/GuWCL18,bulte-tezcan-2019-neural,xu-etal-2020-boosting,cai-etal-2021-neural,hoang-etal-2023-improving,hao-etal-2023-rethinking}, 
where retrieval relies on textual feature matching such as $n$-gram overlap.
These methods are therefore not readily applicable to direct ST due to the continuous nature of speech and much longer input lengths.
In ST, 
\citet{du-etal-2022-non} use $k$NN-MT \cite{DBLP:conf/iclr/KhandelwalFJZL21} for domain adaption. 
This approach requires a joint model for speech and text input, with a fully text-based datastore. 
Our work does not require modifying the ST model to support speech and text inputs, and enables the retriever to query from speech to speech or text. 
Our retrieval module is related to the recent work by \citet{speechDPR} as both are based on DPR.
The main difference is that their model is for question answering and does not support cross-modal retrieval.
\citet{salm} show that LLMs adapted for speech could leverage in-context examples for speech recognition and translation.
Our work is orthogonal to theirs in that we show that conventional encoder-decoder ST models can be trained to exhibit in-context learning abilities.

\paragraph{Rare Words in ASR, MT, and direct ST}
In \textbf{ASR}, 
some representative approaches to handle rare words include language model rescoring or fusion \cite{raju19_interspeech,DBLP:conf/asru/YangLGGRFB21,huang22j_interspeech,weiran22_interspeech,mathur-etal-2023-personalm},
data augmentation by text-to-speech (TTS) \cite{8683745,9414778,DBLP:journals/nn/QuWW23},
and context enhancement by an additional memory module
\cite{8682441,jain20_interspeech,DBLP:conf/asru/ChangLRMORK21,DBLP:conf/asru/HuberHSW21,DBLP:conf/slt/QiuMHS22,DBLP:journals/corr/abs-2401-04482}.
In \textbf{MT}, rare word translation has been tackled by, among other techniques,
constrained decoding \cite{chatterjee-etal-2017-guiding,hasler-etal-2018-neural,ailem-etal-2021-encouraging,zhang-etal-2023-understanding},
copying by source annotations \cite{dinu-etal-2019-training,song-etal-2019-code,bergmanis-pinnis-2021-facilitating} or pointing mechanisms \cite{gulcehre-etal-2016-pointing,pham-etal-2018-towards,gu2019pointer,zhang-etal-2021-point},
and retrieval-augmented translation \cite{martins-etal-2023-empirical,liu-etal-2023-kits}.
In \textbf{direct ST}, translating rare words is a significant challenge due to the combined complexities of ASR and MT. 
The amount of prior work is also relatively sparse.
\citet{gaido-etal-2022-talking} use multilingual models to improve the accuracy of non-English names.
\citet{10094689} propose to first detect named entities (NEs) in the source audio
that are present in a given contextual dictionary 
and then inject these NEs in text form into the decoder.
Our approach does not assume a readily available contextual dictionary, but can instead leverage unprocessed parallel data.

\section{Conclusion}
We introduced a retrieval-and-demonstration approach to improve rare word translation accuracy in direct ST. For real-world applications, e.g., translating scientific talks, we recommend adding utterances from the same speaker to the example pool and using speech-to-speech retrieval to identify examples.
When feasible, 
one should consider incorporating an additional verification step to ensure the relevance of the retrieved sentences, by human-in-the-loop or automated techniques.

\section*{Limitations}

\paragraph{Robustness to Irrelevant Examples}
Our approach effectively improves the accuracy of rare word translation. 
However, as elaborated in the result discussions, we also observed that incorrectly retrieved examples tend to harm translation quality. 
As a next step, we hope to increase the robustness of the ST models to irrelevant examples. 
This could for instance be achieved by incorporating incorrect rare words during training to enhance the model's resilience to such errors.

\paragraph{Targeted Solution for Rare Word Translation}
Our approach is a targeted solution for the use-case of rare word translation.
When there is no rare word in the test sentence, the examples will harm translation quality, as seen in the case of using irrelevant examples. 
Whether rare words exist in the test sentences could be determined by ST model confidence (decoding probability) or retriever distances to the closest neighbor in the example pool.
We leave this exploration to future work.

\paragraph{Language Coverage}
Our experiments were limited to the English-to-German language pair
due to resource constraints.
Experiments on additional language pairs, especially distant ones, would further substantiate the findings.

\paragraph{Extension to other Audio Tasks}
This work focused on rare words in direct speech translation. 
An extension to other audio tasks would enlarge the impact of the proposed approach.
As a partial remedy, we performed preliminary experiments on rare word ASR in \autoref{sec:asr_results} and found that the results support the main findings in this work.

\section*{Acknowledgments}
We thank the anonymous reviewers for their insightful feedback.
We also thank \citet{papi-etal-2024-good} for reporting Conformer bugs which led to unexplainable results in our initial experiments.
Part of this work was performed on the HoreKa supercomputer funded by the
Ministry of Science, Research and the Arts Baden-Württemberg and by
the Federal Ministry of Education and Research.
Part of this work was supported by funding from the pilot program Core-Informatics of the Helmholtz Association (HGF).
Part of this work received support from
the European Union’s Horizon research and innovation programme under grant agreement No
101135798, project Meetween (My Personal AI Mediator for Virtual MEETtings BetWEEN People). 
\bibliography{custom,anthology}

\appendix

\section{Details on Masked Loss} 
\label{appendix:masked_loss}
During the training of our adapted ST model, example sentences are prepended to sentences in the reduced training set. 
The translation of the example sentence is used as a prefix and masked during loss calculation. The cross-entropy loss function we use for training can be expressed as \autoref{eq:L}:
\begin{equation}
  \label{eq:L}
  \mathcal{L}=-\sum_{t=1}^{T} M_t logP(y_t |y_{<t},u^e,y^e,u)
\end{equation}

With $M_t$ as a mask function \autoref{eq:M_t}:
\begin{equation}
    \label{eq:M_t}
    M_t =
    \begin{cases}
      0 & \text{if position $t$ is part of $y^e$}\\
      1 & \text{if position $t$ is part of $y$}\\
    \end{cases}
\end{equation}

\section{Details of Rare Word Types} \label{appendix:rare_word_types}

The detailed rare word analysis results for \autoref{tab:NER analysis} are in \autoref{tab:detailed NER analysis}.

\begin{table}[ht!]
    \small
    \centering
    \setlength\tabcolsep{4pt}
    \begin{tabular}{m{4cm} c} 
    \toprule
    \textbf{Rare Word Type}
    &\textbf{Frequency}\\
    \midrule
    Person &130\\
    Location &72\\
    Technology &29 \\
    Food &27 \\
    Company &25\\
    Biology &23\\
    Organization &18\\
    Health &18 \\
    Culture &14 \\
    Transport &14 \\
    Religion &14 \\
    Fashion &13 \\
    Medicine &12 \\
    Science &12 \\
    Geography &11 \\
    Chemics &11 \\
    Language &11 \\
    History &10 \\
    Politics &9 \\
    Architecture &9 \\
    Military &9 \\
    Environment &8 \\
    Education &7 \\
    Sport &7 \\
    Law &6 \\
    Society &4 \\
    Data &4 \\
    Book &4 \\
    Physics &4 \\
    Game &3 \\
    Economy &3 \\
    Literature &2 \\
    Art &2 \\
    Music &1 \\
    Entertainment &1\\
    Award &1 \\
    \bottomrule
    \end{tabular}
    \caption{Detailed NER results on rare words in tst-rare-word with the number of unique words in each category.}
    \label{tab:detailed NER analysis}
\end{table}

\section{ST Training and Inference Details}\label{appendix:st_training_inference}

\subsection{Training Details}
We use the Transformer architecture \textsc{s2t\_transformer\_s} in \textsc{FairSeq S2T} \cite{wang-etal-2020-fairseq} For all our ST models, the encoder-decoder architecture consists of 12 transformer encoder blocks and 6 transformer decoder blocks, with a model dimension of 256 and an inner dimension (FFN) of 2,048. 

We initialized the ST model from a pre-trained ASR model\footnote{\url{https://dl.fbaipublicfiles.com/fairseq/s2t/mustc_de_asr_transformer_s.pt}}. Subsequently, we fine-tuned the pre-trained model for the ST task with hyperparameters following \cite{wang-etal-2020-fairseq}, specifically, we set dropout rate 0.1 and label smoothing 0.1. The ST training used a tokenizer with a vocabulary size of 8,000. To prevent the tokenizer from seeing the rare words during its  training,
which will cause an unfair test condition,
we train the SentencePiece \cite{kudo-richardson-2018-sentencepiece} tokenizer on the reduced train set after the utterances containing rare words are moved to other splits as discussed in \S\ref{sec:Data Construction}. 

During the training of the adapted ST model with examples, we doubled the effective batch size to maintain a comparable loss scale since the prefix tokens do not contribute to the overall loss. Additionally, we set dropout rate to 0.2 after doing a search in \{0.1, 0.2, 0.3\} based on the dev loss during the training of the adapted ST model. The training was stopped after the validation performance did not improve for 30 consecutive epochs (patience 30). For evaluation, we averaged the last 10 checkpoints.

\subsection{Inference Details}
The inference uses a beam size of 5. 
Since the rare-word-tst dataset includes example-prepended sentences, the sentences are longer than typical translation sentences. To keep all utterances in the rare-word-tst set, we set a large allowed source size with --max-source-positions 30000. 
This ensures that even the longest utterances are not excluded from the rare-word-tst set.
\section{Retriever Training and Inference Details}
\label{appendix:retrieval_training_inference}
\subsection{Training Details}
Our retriever is based on the DPR \cite{karpukhin-etal-2020-dense} architecture, where a dense passage encoder $E_P$ and a question encoder $E_Q$ is constructed to map candidate input $c$ and query input $q$ to latent representation vectors respectively. The similarity between the candidate representation and the query representation is defined as the dot-product of their vectors as shown in \autoref{eq:similarity}:
\begin{equation}
    \label{eq:similarity}
    sim(q,c)=E_Q(q)^T E_P(c)
\end{equation}
The encoders $E_P$ and $E_Q$ of DPR are initialized with SpeechT5 encoder\cite{ao-etal-2022-speecht5} or SONAR encoder \cite{duquenne2023sonar}.
\paragraph{Speech T5} 
The SpeechT5 speech/text encoder transforms speech or text input into a 768-dimensional embedding vector. It comprises 12 Transformer encoder blocks, each with a model dimension of 768 and an inner feed-forward network (FFN) dimension of 3,072. Before the encoder, a speech/text-encoder pre-net preprocesses the input. The speech-encoder pre-net includes the convolutional feature extractor of wav2vec \cite{baevski2020wav2vec} for waveform downsampling. The text-encoder pre-net applies positional encoding to convert character-level tokenized indices into embedding vectors.

\paragraph{SONAR} The SONAR speech/text encoder encodes speech/text input to an embedding vector of 1,024. The encoder consists of 24 transformer encoder blocks with a model dimension of 1,024 and an inner dimension (FFN) of 8,192. The speech encoder-frontend applies the wav2vec feature extractor \cite{baevski2020wav2vec}, while the text encoder-frontend uses a position encoder.

\paragraph{Training}
The dual encoders in DPR are trained on a reduced training set with prepended examples. 
Each sentence’s example works as a positive example, while examples from other sentences in the batch serve as in-batch negatives. We set a batch size of 4 and a learning rate of 2e-5 for training.

Given the large size of the SONAR encoder, for memory efficiency, only the top layer of the encoder is trained. This approach is not only for memory efficiency but also because the lower layers likely extract low-level acoustic features, which are less relevant for our retrieval task focused on word-level information. We further investigate the retrieval accuracy under different numbers of trainable parameters. As shown in \autoref{fig:retrieve_experiments}. We use the settings with the best retrieval accuracy for our ST task. which are: 
\begin{itemize}
\item For the speech-to-speech retriever, the top 2 layers of both speech encoders are trained, resulting in 205 million trainable parameters.
\item For the speech-to-text retriever, the top 8 layers of both the text and speech encoders are trained, with 422 million trainable parameters.
\item For the text-to-text retriever, the top 8 layers of both text encoders are trainable, totaling 335 million trainable parameters.
\end{itemize}

\subsection{Inference Details}
During inference time, we apply the passage encoder $E_P$ to all the candidates in the rare-word pool. Given a question $q$, we can derive its embedding $v_q = E_Q(q)$ and then retrieve the top-1 candidate whose embedding is the closest to $v_q$ from the rare-word pool.

\section{Comparison to Existing Results}
\label{appendix:comparison}

We confirm that our baseline model performs on par with those reported in the literature with the results in \autoref{tab:comparison_fairseq}.

\begin{table}[ht!]
    \small
    \centering
    \begin{tabular}{l c c c c c} 
    \toprule
    &\textbf{BLEU} \\
    \midrule
    \textsc{Fairseq S2T} \cite{wang-etal-2020-fairseq}
    & 22.7  \\
    Our baseline model
    & 23.6 \\
    \bottomrule
    \end{tabular}
    \caption{The performance of our baseline model on the tst-COMMON split of MuST-C is comparable to existing baselines. 
    Both models have the identical architecture using \textsc{s2t\_transformer\_s}. }
    \label{tab:comparison_fairseq}
\end{table}    

\section{Additional Examples}
\label{appendix:sentence_examples}
Here we present two additional translation examples for comparison among the baseline model, the model trained with an additional rare-word pool, and our approach. In the first example, our approach successfully translates a zero-shot word perfectly. In the second example, we demonstrate a case where our approach does not perform well.
\begin{table}[t]
	\centering
	\small 
	\begin{tabularx}{\columnwidth}{X}
	\toprule
		\textbf{source} (transcript): {\color{violet}Murali Krishna} {\color{gray}(Murali Krishna)} comes from one of those villages.
        \\
		\textbf{baseline model (on train-reduced)} (\autoref{tab:rareword_translation_accuracy} row $(1)$):{\color{purple}Moralische Christen} {\color{gray}(Moral Christians)}  sind aus einem dieser Dörfer.
	\\
        \textbf{train on \{train-reduced + rare-word pool\}} (\autoref{tab:rareword_translation_accuracy} row $(4)$):
        Das {\color{purple}Marate Krishna} {\color{gray}(Marate Krishna)} kommt aus einem dieser Dörfer.
        \\
        \textbf{speech$\rightarrow$speech example} (\autoref{tab:retrieval_performance} row $(5)$): Sie arbeitet mit Leuten wie {\color{blue}Murali Krishna}. {\color{gray}(She works with people like Murali Krishna.)}.
        \\
        \textbf{adapted + speech$\rightarrow$speech} (\autoref{tab:rareword_translation_accuracy} row $(7)$): 
        {\color{blue}Murali Krishna} {\color{gray}(Murali Krishna)} kommt aus einem dieser Dörfer.
            \\
        \textbf{target}:
         {\color{blue}Murali Krishna} {\color{gray}(Murali Krishna)} kommt aus einer dieser Dörfer.
         \\
        \midrule
        \textbf{source} (transcript): The {\color{violet}McLaren} {\color{gray}(McLaren)} just popped off and scratched the side panel.
            \\
        \textbf{baseline model (on train-reduced)} (\autoref{tab:rareword_translation_accuracy} row $(1)$):Und der {\color{purple}Klient} {\color{gray}(client)} stoppte ab und kratzte die Seite des Paddels.
	\\
        \textbf{train on \{train-reduced + rare-word pool\}} (\autoref{tab:rareword_translation_accuracy} row $(4)$): Und der {\color{purple}Spieler} {\color{gray}(player)} stürzte einfach ab und kratzte auf den Bürgersteig.
         \\
         \textbf{speech$\rightarrow$speech example} (\autoref{tab:retrieval_performance} row $(5)$): Aber als Nebeneffekt sammelt er Kornette. {\color{gray}(But as a sideline, he happens to collect cornets.)}
            \\
        \textbf{adapted + speech$\rightarrow$speech} (\autoref{tab:rareword_translation_accuracy} row $(7)$): Als der {\color{purple}Klairner} {\color{gray}(Klairner)} gerade ankam, stopfte er ein Nebenpandel.
            \\
        \textbf{target}: Der {\color{blue}McLaren} {\color{gray}(McLaren)} bekam eine Beule und einen Kratzer an der Seitenkarosserie.
        \\
	\bottomrule
	\end{tabularx}	
	\caption{Additional examples of our retrieval-and-demonstration approach.}
        \label{tab:additional_example}
\end{table}

\section{Preliminary ASR Results}
\label{sec:asr_results}
To test the generalizability of our approach, 
we additionally ran rare word ASR experiments on the same data splits following the data construction steps in \S\ref{sec:Data Construction}.
The results are in \autoref{tab:asr}.
Here we directly used all hyperparameters for the ST models. 
The scores therefore may be not optimal.
However, pir main findings still hold given the additional results:
\begin{enumerate}
\item ASR models can also effectively learn from demonstration at inference time: Rare word recognition accuracy in line (2) vs. (1) improves from 31.2 to 72.1\%.
\item Seeing rare words only in training does not sufficiently improve their recognition accuracy: Rare word accuracy does not improve as much in line (4) vs. (1) compared to (2) vs. (1).
\item Speech→speech outperforms speech→text retrieval: In systems with retrieved examples, line (7) has the best performance.
\end{enumerate}

\begin{table*}[t]
    \small
    \centering
    \begin{tabular}{l c c c c c}  
    \toprule
    \textbf{ASR Model} 
    & \textbf{WER} 
    & \makecell{\textbf{Overall} \\ \textbf{acc (\%)}} 
    & \makecell{\textbf{0-shot} \\ \textbf{acc (\%)}} 
    & \makecell{\textbf{1-shot} \\ \textbf{acc (\%)}} \\
    
    \midrule
    (1) baseline model (on train-reduced)
    &14.8 & 31.2 & 27.0  & 40.3  \\
    (2) adapted + gold example
    &22.0  &\textbf{72.1} &\textbf{71.4}  &\textbf{73.8}  \\
    (3) adapted + random example
    &25.3 & 19.8 & 18.6 & 22.4 \\
    (4) train on \{train-reduced + rare-word pool\} (more data)
    &\textbf{13.9} &42.8 &38.7  &51.7  \\
    \midrule
    \textbf{Using retrieved examples} \\
    (5) adapted + text (gold transcript)$\rightarrow$text 
    &28.0 &46.2 &45.0  &\textbf{48.8} \\
    (6) adapted + speech$\rightarrow$text
    &28.1 &40.1 &39.3  &41.7 \\
    (7) adapted + speech$\rightarrow$speech
    &\textbf{21.7} &\textbf{46.8} &\textbf{46.2}  &48.1 \\
    \bottomrule
    \end{tabular}
    
    \caption{
    ASR quality (WER$\downarrow$) and rare word accuracy$\uparrow$ (overall, 0- and 1-shot) of different models on the tst-rare-word split. 
    The lower section uses retrieved examples from the retriever (\S\ref{subsec:result_st_with_retrieval}).
    }
    \label{tab:asr}
\end{table*}

\end{document}